\def\year{2022}\relax
\documentclass[letterpaper]{article} 
\usepackage{aaai22}  
\usepackage{times}  
\usepackage{helvet}  
\usepackage{courier}  
\usepackage[hyphens]{url}  
\usepackage{graphicx} 
\usepackage{natbib}  
\usepackage{caption} 
\DeclareCaptionStyle{ruled}{labelfont=normalfont,labelsep=colon,strut=off} 
\usepackage{algorithm}
\usepackage{algorithmic}
\usepackage{diagbox}
\usepackage{amsmath}
\usepackage{amssymb}
\usepackage{newfloat}
\usepackage{listings}
\floatstyle{ruled}
\newfloat{listing}{tb}{lst}{}
\floatname{listing}{Listing}
\title{Rethinking the Knowledge Distillation From the Perspective of Model Calibration}
\author{
    Lehan Yang,
    Jincen Song\\
}
\affiliations{
    School of Computer Science, The University of Sydney, NSW, Australia\\

    \{lyan3310, json4256\}@uni.sydney.edu.au
}

\usepackage{bibentry}

\begin{document}

\maketitle

\begin{abstract}
Recent years have witnessed dramatically improvements in the knowledge distillation, which can generate a compact student model for better efficiency while retaining the model effectiveness of the teacher model.
Previous studies find that: more accurate teachers do not necessary make for better teachers due to the mismatch of abilities. In this paper, we aim to analysis the phenomenon from the perspective of model calibration. We found that the larger teacher model may be too over-confident, 
thus the student model cannot effectively imitate. While, after the simple model calibration of the teacher model, the size of the teacher model has a positive correlation with the performance of the student model.
\end{abstract}

\section{Introduction}
Knowledge distillation \cite{hinton2015distilling} is widely used to compress the deep neural networks, which employs the soft probabilities of a lager teacher network to guide the training of smaller student network \cite{cho2019efficacy}. The applications of knowledge distillation seem to be evident, such as model compression, transfer learning, domain adaption, incremental learning. Many follow-ups work endeavor to improve the performance of knowledge distillation, using different strategies. Despite sustainable efforts have been made, many studies \cite{mirzadeh2020improved} suggest that: more accurate teacher models do not necessarily make for better teachers. The small model does not imitate the large model very well, when the performance gap between the large and small models is large. Researchers attributed the phenomena to the mismatch of abilities between the teacher and student. What characterizes this phenomenon? Are the teacher models over-confident? In our study, we aim to answer these questions. 

\section{Methodology}

The overall pipeline of our framework is shown in Fig. 1, which consists of two models of CNN architectures (the teacher model and the student model). 
The main between difference between our work and vanilla knowledge distillation is whether the model calibration is used, while the motivation of this work is to investigate whether the model calibration can be helpful to address aforementioned performance decrease of the student model. We will explain the components in more details subsequently.

\subsection{Knowledge Distillation}
As aforementioned, knowledge distillation can minimize the distance between teacher's and student's output, and completing the process of knowledge transfer.
\cite{cho2019efficacy,mirzadeh2020improved} have observed that: when the size of the teacher model increases, the accuracy of the student model may decrease if the performance gap between the teacher and the student is large. They speculated that it was due to the mismatch of the capacity, as the tiny student model cannot imitate the large teacher model well. In our experiments, we also reproduced the experiment results and observed similar phenomenon.
In the vanilla knowledge distillation, we find that the large-scale teacher model is not necessarily better than the small-scale teacher model, as shown in the experimental sections.

\subsection{Model Calibration} 
The prediction results of the deep model, such as ResNet, are becoming more over-confidence and the output of the model cannot accurately represent the confidence of classification probability. \cite{guo2017calibration} observed that the model depth width, weight decay and batch normalization are the important factors that affect model calibration. To address the over-confident issue, one simple solution is to minimize the negative log likelihood (NLL), which introduce as Eq. \ref{nll}, to train the temperature parameter $T$, and use the temperature scale method to perform simple post-processing calibration on the model. As the softmax function after temperature scaling shows in Eq. \ref{ts}, $\mathbf{z}_{i}$ donate the logit vector from last layer of network, $T$ is the temperature parameter optimized, $\hat{q}_{i}$ donate the calibrated confidence.

\begin{equation} \label{nll}
\mathcal{L}=-\sum_{i=1}^{n} \log \left(\hat{\pi}\left(y_{i} \mid \mathbf{x}_{i}\right)\right)
\end{equation}

\begin{equation} \label{ts}
\hat{q}_{i}=\max _{k} \quad \sigma_{\mathrm{SM}}\left(\mathbf{z}_{i} / T\right)^{(k)}
\end{equation}

In the calibration of the teacher model, we can just simply change the original knowledge temperature $t$ to the optimal temperature scaling temperature $T$, as the knowledge distillation softmax layer equation shows in Eq. \ref{kd_cal}. To evaluate the performance of model calibration, the Expected Calibration Error (ECE) is used in our experiments, whose definition is given in Eq. \ref{ece}.

\begin{equation} \label{kd_cal}
y_{i}(\mathbf{x} \mid t)=\frac{e^{\frac{z_{i}(\mathbf{x})}{t}}}{\sum_{j} e^{\frac{z_{j}(\mathbf{x})}{t}}} \longrightarrow \frac{e^{\frac{z_{i}(\mathbf{x})}{T}}}{\sum_{j} e^{\frac{z_{j}(\mathbf{x})}{T}}}
\end{equation}

\begin{equation} \label{ece}
\mathrm{ECE}=\sum_{m=1}^{M} \frac{\left|B_{m}\right|}{n}\left|\operatorname{acc}\left(B_{m}\right)-\operatorname{conf}\left(B_{m}\right)\right|
\end{equation}

\begin{figure}[t]
\centering
\includegraphics[width=1\columnwidth]{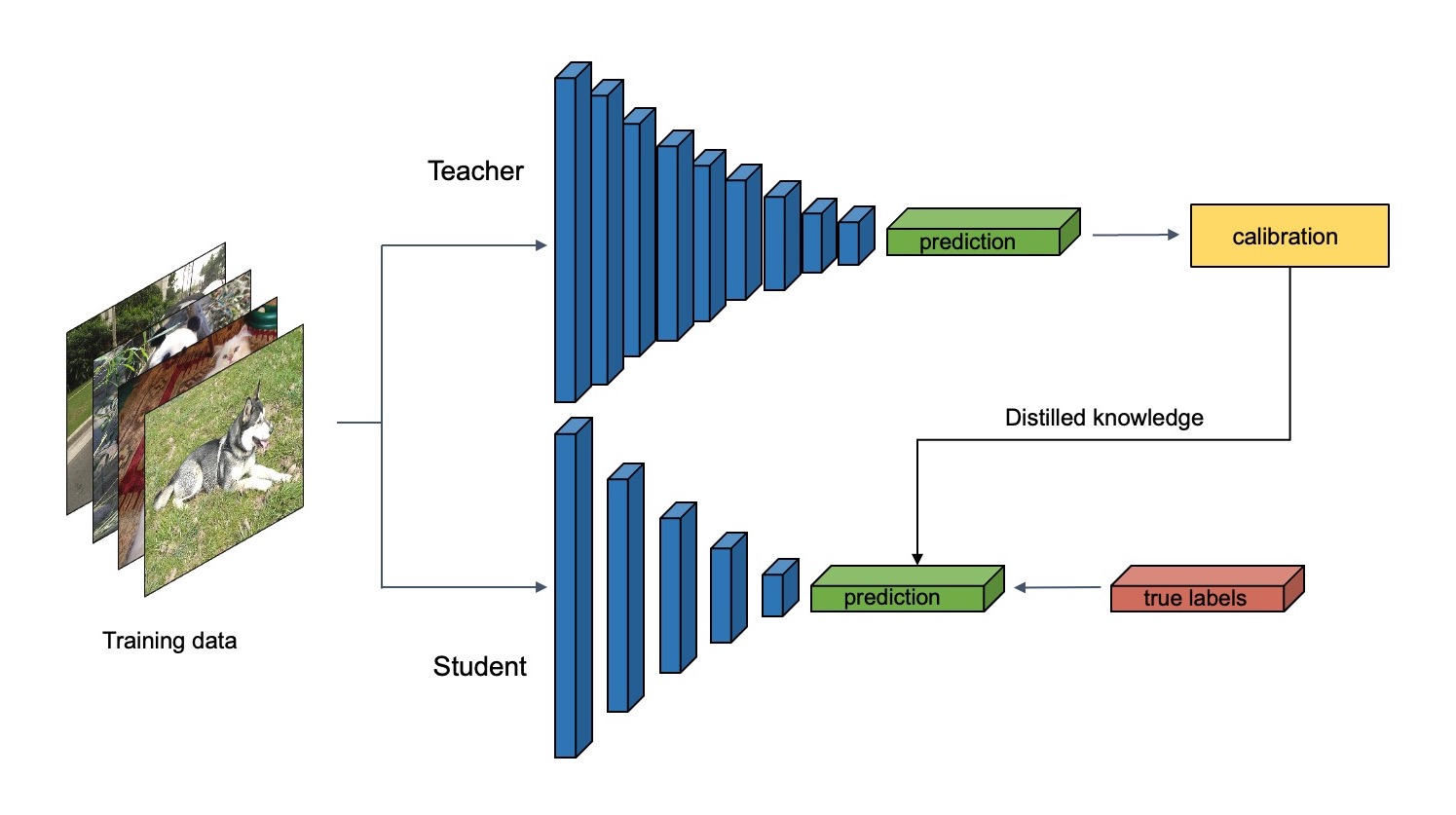} 
\caption{The framework of our calibrated knowledge distillation.}
\label{framework}
\end{figure}




\section{Experiments Results}
\subsection{Experimental setup}
In our experiments, we use CIFAR-10 and Fashion-MNIST as the tony examples. The batch size is set as 128, while the cosine annealing learning rate and the initial rate is set as 0.1. For the teacher model, we use the ResNet and the SGD optimizer. The maximum epoch is set as 200. For the vanilla Knowledge Distillation, we use the same parameters as the training teacher model, set the $\alpha$ to 0.8. Then, we minimize the NLL on the validation set to obtain the optimal temperature parameters, use the temperature scaling method to correct the output of the teacher model.

\begin{table}[t]
\begin{center}
\caption{\label{tab:baseline} This table shows that the classification accuracy of ResNet baseline on CIFAR-10 dataset.}
\begin{tabular}{ccccc}
\hline
ResNet18 & 34 & 50 & 101 & 152 \\
\hline

0.9542 & 0.9565 & 0.9553 & \textbf{0.9567} & 0.9557\\
\hline
\end{tabular}
\end{center}
\end{table}

\begin{table}[t]
\begin{center}
\caption{\label{tab:temp_scale} The ECE and NLL before and after the temperature scale calibration applied on ResNet teacher network.}
\begin{tabular}{c|cccc}
\hline
 & 34 & 50 & 101 & 152 \\
 \hline
Optimal Temp  & 1.285 & 1.393 & 1.386 & 1.473\\
\hline
ECE Before & 0.035 & 0.031 & 0.034 & 0.033 \\
\hline
ECE After & \textbf{0.020} & \textbf{0.014} & \textbf{0.021} & \textbf{0.015} \\
\hline
NLL Before & 0.240 & 0.255 & 0.274 & 0.273 \\
\hline
NLL After & \textbf{0.214} & \textbf{0.228} & \textbf{0.248} & \textbf{0.239} \\
\hline
\end{tabular}
\end{center}
\end{table}

\begin{table}[t]
\begin{center}
\caption{\label{tab:cali_compare} The classification accuracy of ResNet using vanilla knowledge distillation on CIFAR-10 dataset. The integers represent the depth of ResNet, separated by calibrated or not.}
\begin{tabular}{c|cccc}
\hline
\diagbox{T}{S}& 18 & 34 & 50 & 101  \\
\hline
50 & 0.9561  & \textbf{0.9589}& - & - \\
Calibrated 50 &  \textbf{0.9564} & 0.9559& - & - \\
\hline
101 & 0.9539  & 0.9557 & 0.9561 &  - \\
Calibrated 101 & \textbf{0.9554} &  \textbf{0.9569}& \textbf{0.9591} &  - \\
\hline
152 & 0.9541 & 0.9569  & 0.9551 & 0.9564 \\
Calibrated 152 & \textbf{0.9551}&\textbf{ 0.9577}& \textbf{0.9593} &\textbf{ 0.9592} \\

\hline
\end{tabular}
\end{center}
\end{table}

\subsection{Results}

Comparing with the accuracy in Table \ref{tab:baseline} and the uncalibrated vanilla knowledge distillation in Table \ref{tab:cali_compare}, it can be seen that better teacher does not necessarily make a better student, and the performance of the student model will decrease.

Through the temperature scaling, the calibration is conducted on the teacher model. It can be seen in Table \ref{tab:cali_compare} that both the ECE and NLL have dropped after calibration, which means that the model has been successfully calibrated. After the calibration, most of the accuracy of the student model has been improved. Moreover, after calibration, more large models can bring better performance than small teacher models. We also did experiments on Fashion-MNIST dataset, and observed similar phenomenon. Due to the space constraint, the detailed results are not presented here.

\section{Conclusion}
In this abstract, we explore to analysis the knowledge distillation from the perspective of model calibration. We suppose: the phenomenon that a large model is not necessarily a good teacher could be due to the over-confidence of the large model. Primary results suppose our hypothesis, using the CIFAR-10 and Fashion-MNIST datasets.
 
\bibliography{main.bib}

\end{document}